\documentclass[review]{elsarticle}

\usepackage{lineno,hyperref}
\modulolinenumbers[5]
\usepackage{amssymb}
\usepackage{graphicx}
\usepackage{amsmath}
\usepackage{enumerate}
\usepackage{bbding} 

\usepackage{tabularx,booktabs}
\usepackage{array}
\usepackage{algorithm}
\usepackage{algorithmic}
\usepackage{url}
\usepackage{float}  
\usepackage{subfigure}
\usepackage{multirow}  

\usepackage{lipsum}
\makeatletter
\newenvironment{breakablealgorithm}
{
	\begin{center}
		\refstepcounter{algorithm}
		\hrule height.8pt depth0pt \kern2pt
		\renewcommand{\caption}[2][\relax]{
			{\raggedright\textbf{\ALG@name~\thealgorithm} ##2\par}%
			\ifx\relax##1\relax 
			\addcontentsline{loa}{algorithm}{\protect\numberline{\thealgorithm}##2}%
			\else 
			\addcontentsline{loa}{algorithm}{\protect\numberline{\thealgorithm}##1}%
			\fi
			\kern2pt\hrule\kern2pt
		}
	}{
		\kern2pt\hrule\relax
	\end{center}
}
\makeatother
\newcommand{\tabincell}[2]{
	\begin{tabular}{@{}#1@{}}#2\end{tabular}
}

\journal{Applied mathematical modeling}

\begin{document}
\begin{frontmatter}
	
	\title{The Capacity Constraint Physarum Solver}
	
	
	\author[l1]{Yusheng Huang}
	\author[l2]{Dong Chu}
	\author[l1]{Yong Deng\corref{cor1}}
	\ead{dengentropy@uestc.edu.cn, prof.deng@hotmail.com}
	\author[l3,l4]{Kang Hao Cheong\corref{cor1}}
	\ead{kanghao\_cheong@sutd.edu.sg}
	
	\address[l1]{Institute of Fundamental and Frontier Science, University of Electronic Science and Technology of China, Chengdu, 610054, China}
	\address[l2]{Schools of Information and Communication Engineering, University of Electronic Science and Technology of China, Chengdu, 610054, China}
	\address[l3]{Science, Mathematics and Technology Cluster, Singapore University of Technology and Design (SUTD), S487372, Singapore}
	\address[l4]{SUTD-Massachusetts Institute of Technology International Design Centre, Singapore}

	\cortext[cor2]{Corresponding author.}
	\begin{abstract}
	Physarum polycephalum inspired algorithm (PPA), also known as the Physarum Solver, has attracted great attention. By modelling real-world problems into a graph with network flow and adopting proper equations to calculate the distance between the nodes in the graph, PPA could be used to solve system optimization problems or user equilibrium problems. However, some problems such as the maximum flow (MF) problem, minimum-cost-maximum-flow (MCMF) problem, and link-capacitated traffic assignment problem (CTAP), require the flow flowing through links to follow capacity constraints. Motivated by the lack of related PPA-based research, a novel framework, the capacitated physarum polycephalum inspired algorithm (CPPA), is proposed to allow capacity constraints toward link flow in the PPA. To prove the validity of the CPPA, we developed three applications of the CPPA, i.e., the CPPA for the MF problem (CPPA-MF), the CPPA for the MCFC problem, and the CPPA for the link-capacitated traffic assignment problem (CPPA-CTAP). In the experiments, all the applications of the CPPA solve the problems successfully. Some of them demonstrate efficiency compared to the baseline algorithms. The experimental results prove the validation of using the CPPA framework to control link flow in the PPA is valid. The CPPA is also very robust and easy to implement since it could be successfully applied in three different scenarios. The proposed method shows that: having the ability to control the maximum among flow flowing through links in the PPA, the CPPA could tackle more complex real-world problems in the future.  
	\end{abstract}
	
	
	
	
\end{frontmatter}

\section{Introduction}
\label{sec.1}
The bio-inspired algorithm (BIA) is recently a heated topic. Various kinds of BIAs, such as differential evolution algorithm\cite{A_DE}, genetic algorithm \cite{A_GA}, ant colony algorithm (ACO)\cite{A_ACO}, and artificial bee colony algorithm \cite{A_ABC} have been developed. The BIAs have demonstrated their success in solving complex problems, such as global numerical optimization \cite{A_example1}, numerical function optimization \cite{A_ABC}, parameter estimation \cite{A_example2}, just to name a few. Thus, many researchers have dedicated to develop new BIAs, enhance the existing BIAs, and expand the areas of application for current BIAs.

In recent years, a newly developed BIA named the Physarum polycephalum inspired algorithm (PPA) \cite{A_orginalPPA}, also called the physarum solver, has attracted great attention. Physarum polycephalym is an acellular slime mold or myxomycete, which is a living substrate and has a structure similar to a network of tubes \cite{A_PPAmaze}. When foraging, it will form a network of tubes that connects its growth point and food sources. The network-liked structure of the physarum polycephalym's body allows nutrients and chemical signals to circulate throughout the organism \cite{A_PPAmaze}. When stimulated by the external factors, physarum polycephalum would adjust the thickness of the tubes and even the self-structure as a response \cite{GG_xiaogeTAP}. What attracts the researchers' attention is that the forming of the tubes of the physarum polycephalym has a positive feedback mechanism: if the flux of protoplasm shuttled through a certain tube persists or increases for a certain time, the tube would become thicker; otherwise, the tube would become thinner \cite{GG_IPPA}. The feedback mechanism implies the path-finding ability of the physarum polycephalym. Inspired by this interesting mechanism, Tero \textit{et al.} \cite{A_orginalPPA} constructed a mathematical model (the original PPA) to describe the path-finding behavior of the physarum polycephalym. Later, Tero \textit{et al.} demonstrated that the original PPA could solve mazes \cite{A_PPAmaze} and even find the shortest path in a certain given graph \cite{A_orginalPPA}. After that, Bonifaci \textit{et al.} mathematically proved the convergence of the original PPA on the shortest path problems \cite{A_prove2} and calculated time complexity bounds \cite{A_prove1}. 

Recent studies of the PPA could be divided into two parts, i.e., one is to improve the performance of the original PPA, and the other aims to expand the applications of the PPA. Among the former kind of researches, researchers mainly focus on developing new adaptation equations or new termination criteria to enhance the performance of the original PPA. For example, Zhang \textit{et al.} \cite{GG_IPPA} accelerated the original PPA in the shortest path problem by introducing the concept of conservation of energy into it; Chao \textit{et al.} \cite{A_PPAEx1} fastened the original PPA by pruning inactive nodes and terminating the physarum solver in advance using new termination criteria; Wang \textit{et al.} \cite{A_PPAEx2} proposed an anticipation mechanism to allow the original PPA to converge much faster. Some researchers have made their efforts in expanding the applications of the PPA. For example, Yang \textit{et al.} \cite{A_PPAEx3} developed the PPA for fuzzy user equilibrium; Liu \textit{et al.} \cite{A_PPAEx4} proposed a PPA-based algorithm for the steiner tree problem in networks; Guo \textit{et al.} \cite{A_PPAEx5} constructed a physarum-inspired obstacle-avoiding routing algorithm.

After reviewing the previous studies of the PPA, we found that a few studies enabled the capacity constraints on the tubes in the PPA. Current PPA's applications are mainly about system optimization problems and user equilibrium problems that do not contain capacity constraints. For example, Zhang \textit{et al.} \cite{GG_xiaogeTAP} proposed the PPA for traffic assignment problem and Xu \textit{et al.} \cite{GG_PPATAP} improved its performance in the multiple O-D pairs scenario. But none of them consider the situation that: in the real world, each road has a traffic-flow capacity. As a result, though the above two methods could converge to the equilibrium, the traffic flow in some links exceeds far away from the roads' capacity (Readers could refer to Table \ref{table.resultCTAP} for this phenomenon.). That is not going to be possible in the real world. Thus, embedding capacity constraints into the PPA would be necessary so that we could apply the PPA in some more real-world-liked problems. Wang \textit{et al.} \cite{GG_PPAMC_ZWang} and Wang \cite{GG_PPAMC_QWang} made preliminary attempts in controlling the flow on the link of the PPA when applying the PPA to solve the maximum flow (MF) problem. Wang \cite{GG_PPAMC_QWang} introduced the concept of basic link flow to make the flow flowing through each link comply with the capacity constraint. Wang \textit{et al.} \cite{GG_PPAMC_ZWang} introduced the concept of energy flow to control the flow flowing through each link and proposed a newly developed PPA-liked method. However, both of the above methods are very complex, making it hard to migrate to other applications. Also, the new concepts introduced by them into the PPA for the MF problem affect the pathfinding behavior of the PPA. To be more specific, they changed the adaptation Eq.(\ref{eq.5}), but some applications of the PPA also need special adaptation equation, such as \cite{GG_IPPA}. Thus, changing the adaptation equation might make the PPA not suitable for optimization problems. What's more, Wang's method \cite{GG_PPAMC_ZWang} consumes a high amount of running time even though the graph scale is small and Wang's method \cite{GG_PPAMC_QWang} seems only efficient to handle the directed graph.

Thus, considering the reasons above, we have proposed the capacitated physarum polycephalum inspired algorithm (CPPA) by using a straight forward way to enable the capacity constraints of the link flow in the physarum solver. Three applications of the CPPA, i.e., the CPPA for the MF problem (CPPA-MF), the CPPA for the minimum-cost-maximum-flow problem (CPPA-MCMF), and the CPPA for the link-capacitated traffic assignment problem (CPPA-CTAP), have been developed to demonstrate the effectiveness and robustness of the CPPA. Our contributions are listed below:
\begin{itemize}
	\item We have developed a novel framework, called the CPPA, to control the link flow in the physarum solver, which is easy to implement and migrate to other applications.
	\item An application of the CPPA, the CPPA-MF is provided. Experiments have demonstrated that the CPPA-MF is effective in MF problem solving.
	\item An application of the CPPA, the CPPA-MCMF is provided. Experiments have demonstrated that the CPPA-MCMF is effective in MF problem solving and is faster than the baseline algorithms. It means that the CPPA does not affect the PPA's optimization ability while enabling capacity constraints. What's more, it is \textbf{the first} PPA-based algorithm that could tackle the MCMF problem.  
	\item An application of the CPPA, the CPPA-CTAP is provided. Experiments have demonstrated that the EDPPA-TAP is effective in CTAP problem solving, which means that the CPPA could be applied to a real-world liked problem. What's more, it is \textbf{the first} PPA-based algorithm that could tackle the CTAP problem. 
\end{itemize}

The rest of the paper is organized as follows ection \ref{sec.2} provides some background information on the PPA, the MF problem, the MCMF problem, and the CTAP problem. Section \ref{sec.3} provides the proposed CPPA framework, the proposed CPPA-MF algorithm, the proposed CPPA-MCMF algorithm, and the proposed CTAP algorithm. Our proposed methods are then experimentally tested in section \ref{sec.4}. Section \ref{sec.5} concludes the paper.

\section{Brief introduction of background knowledge}
\label{sec.2}
In this section, we briefly provide some background knowledge. Section \ref{sec.2.1} introduces the mathematical model of the physarum polycephalym; Section \ref{sec.2.2}, Section \ref{sec.2.3} and Section \ref{sec.2.4} provide some basic information on the MF problem, the MCMF problem and the CTAP problem, respectively.
\subsection{The original PPA}
\label{sec.2.1}
In this part, a brief introduction to the original PPA is given. More details can be found in \cite{A_orginalPPA}. Consider a tube-liked network formed by the physarum polycephalum is a graph $G(M,N)$ (except where specifically noted, the 'graph' in the remaining of the paper refers to the undirected or directed graph without rings and negative-weights edges). 

The network's tubes (links) and the junctions between them are abstracted as the edges and nodes in the graph $G(M,N)$ respectively. The growing points of the organism are considered as the starting nodes of the network and the food sources are considered as the ending nodes. For convenience, consider the following specific case: the physarum polycephalum only has one growing point and there is only one point of the food source. For other scenarios, please refer to materials \cite{GG_xiaogeTAP,GG_PPATAP}. The starting node and the ending node are labeled as $N_1$ and $N_2$, respectively. For the other nodes in the graph, they are labeled as $N_3, N_4, N_5, \cdots$. The edge between nodes $N_i$ and $N_j$ is represented by $M_{ij}$. $Q_{ij}$ is the flux from node $N_i$ to node $N_j$ through the edge $M_{ij}$. 

\subsubsection{The flowing pattern in the graph}
\label{sec.2.1.1}
According to \cite{A_orginalPPA}, the flow in this network should fit the definition of Poiseuille flow. Thus, $Q_{ij}$ can be written as \cite{A_orginalPPA}:
\begin{equation}
Q_{ij}=
\frac{D_{ij}}{L_{ij}}(p_i-p_j),
\label{eq.1}
\end{equation}
where $D_{ij}$ and $L_{ij}$ denote the conductivity and the length of the edge $M_{ij}$ respectively while $p_i$ is the pressure at node $N_i$.

Consider the inflow and the outflow of the entire network are equal and constant, which is $I_0$. Then, at the starting and the ending node ($N_1$ and $N_2$) of the graph, the following equation could be derived \cite{A_orginalPPA}:
\begin{equation}
\left \{\begin{gathered}
\sum_{i\not=1}{Q_{i1}+I_0=0}  \\
\sum_{i\not=2}{Q_{i2}-I_0=0}  \\
\end{gathered} \right. .
\label{eq.3}
\end{equation}

At each node except the starting and ending node, the inflow and outflow should be balanced \cite{A_orginalPPA}:
\begin{equation}
\sum_{j\not=1,2; i\not=j}{Q_{ij}=0}.
\label{eq.2}
\end{equation}

Now, combining equations (\ref{eq.1}) to (\ref{eq.3}), the network Poisson equation can be obtained as follows \cite{A_orginalPPA}:
\begin{equation}
\sum_{i}{\frac{D_{ij}}{L_{ij}}(p_i-p_j)=\left \{\begin{gathered}
	+I_0\ \ for\ j=1,  \\
	-I_0\ \ for\ j=2,  \\
	0\ \ \ otherwise.  \\
	\end{gathered} \right.}
\label{eq.4}
\end{equation}

With $p_1$ set to $0$, all $p_i$ can be determined by solving Eq.(\ref{eq.4}). $Q_{ij}$ can be derived from (\ref{eq.1}). In order to solve Eq.(\ref{eq.4}), $D_{ij}$ is needed, which is provided in the next sub-subsection.

\subsubsection{The adaptive behavior of the tube forming}
\label{sec.2.1.2}
In \cite{A_orginalPPA}, the following \textbf{adaptation equation} is used to accommodate the adaptive behavior of the tube forming:
\begin{equation}
\frac{d}{dt}D_{ij}=f(|Q_{ij}|)-rD_{ij},
\label{eq.5}
\end{equation}
where $f(|Q_{ij}|)=|Q_{ij}|$ and $r=1$ are typically adopted. This adaptation equation suggests that the increase or decrease rate of the conductivity of each tube is highly relevant to not only its conductivity but also the current flow flowing through it.

Eq.(\ref{eq.5}) could be written as follows after discretization \cite{A_orginalPPA}:
\begin{equation}
\frac{D_{ij}^{n+1}-D_{ij}^{n}}{\Delta t}=|Q_{ij}^n|-D_{ij}^{n+1}, 
\label{eq.6}
\end{equation}
where $\Delta t$ is usually considered as $1$; $Q_{ij}^n$ represents the flux through the edge $M_{ij}$ at the $n$th iteration; $D_{ij}^{n+1}$ and $D_{ij}^{n}$ denote the conductivity of the edge $M_{ij}$ at the $n+1$th and the $n$th iteration respectively. The above equation could be reformulated as \cite{A_orginalPPA}:
\begin{equation}
D_{ij}^{n+1}=\frac{|Q_{ij}^n|+D_{ij}^n}{2}. 
\label{eq.7}
\end{equation}

\subsubsection{A general procedure of the original PPA}
\label{sec.2.1.3}
A general procedure of the original PPA at $n^{th}$ iteration is written as follows ($L_{ij}$ and $I_0$ are given before the start of iteration) \cite{A_orginalPPA}:
\begin{enumerate}[Step 1]
	\item Input $D_{ij}^{n}$ from last iteration.
	\item With $L_{ij}$ and $I_0$, solve Eq.(\ref{eq.4}) to obtain $p_i$.
	\item With $p_i$, $L_{ij}$ and $D_{ij}^{n}$, solve Eq.(\ref{eq.1}) to obtain $Q_{ij}$.
	\item Generate $D_{ij}^{n+1}$ for the next iteration using $Q_{ij}$ and $D_{ij}^{n}$ according to Eq.(\ref{eq.7}). 
\end{enumerate}

\subsection{The maximum flow (MF) problem}
\label{sec.2.2}
The maximum flow (MF) problem involves finding a maximum feasible flow passing through a flow network. In 1956, Ford and Fulkerson demonstrated the famous max-flow min-cut theorem: the maximum flow from the source to the sink equals the sum of edges in the minimum cut \cite{ford1956maximal}. There are mainly two categories of algorithms for min-cut/max-flow problems: the augmenting paths-based algorithms (such as the Dinic algorithm \cite{dinic1970algorithm}), and the push-relabel algorithms \cite{goldberg1988new}.

In the MF problem, the amount of flow $Q_{ij}$ allowed on each edge $M_{ij}$ is restricted by the capacity $C_{ij}$, which is:
\begin{equation}
0\leq Q_{ij}\leq C_{ij}
\label{eq1}
\end{equation}

\begin{equation}
\label{eq2}
\sum_{M_{ij} \in M} Q_{ij}-\sum_{M_{ji} \in M} Q_{ji}=\left\{\begin{array}{ll}
Q_{st} & i=s \\
0 & i \in N \backslash \{s, t\} \\
-Q_{st} & i=t
\end{array}\right.
\end{equation}
where $N$ and $M$ denotes the set of nodes and edges in the graph respectively; $s$ is the starting node and $t$ is the ending node in the graph.

Thus, the standard MF problem is given as follows:
\begin{equation}
\label{eq3}
\text{max} \quad f_{st} \\
\end{equation}
\begin{equation*}
\text{s.t.}\quad Eq.(\ref{eq1})\quad and \quad Eq.(\ref{eq2})
\end{equation*}

\subsection{The minimum-cost-maximum-flow (MCMF) problem}
\label{sec.2.3}
The minimum-cost maximum-flow (MCMF) problem is a variation of the minimum-cost flow problem. The MCMF problem aims to find a minimum-cost flow among all MF problem's solutions \cite{moolman2020maximum}. Various real-life applications have demonstrated the significance and importance of the MCFC problem. Yang \textit{et al.} modified the traditional MCMF to describe traffic assignment problems in the process of special events evacuation \cite{yang2008evacuation}. Ababei \textit{et al.} reformulated the distribution network reconfiguration problem into a MCMF problem \cite{ababei2010efficient}. In some cases, such as the resource allocation in clouds \cite{hadji2012minimum} and image reconstruction \cite{moisi2013maximum}, the near-optimal solutions could be achieved by solving the corresponding MCMF problems abstracted from the original real-world problems.

The MCMF problem is modelled as:
\begin{equation}
\text{min}\quad \sum_{M_{ij}\in M}cost_{ij}Q_{ij}
\end{equation}
\begin{equation*}
\text{s.t.} \quad \{Q_{ij}\} \in \mathop{\arg\max}_{\{Q_{ij}\}} f_{st}
\end{equation*}
where $cost_{ij}$ is the unit cost of link $M_{ij}$; $\{Q_{ij}\} \in \mathop{\arg\max}_{\{Q_{ij}\}} f_{st}$ means one solution of the MF problem. Thus, the MCMF problem is to find one or more solution which has the minimum cost among all the solutions of the MF problem.

\subsection{The link-capacitated traffic assignment (CTAP) problem}
\label{sec.2.4}
The classic traffic assignment problem (TAP) is that: given an oriented-destination (OD) pair, traffic demand should be assigned to proper traffic routes that connect the OD pair in the traffic network \cite{GG_PPATAP}. How to do the assignment is the core of the TAP. Wardrop \cite{CATP_wardrop} proposed the user equilibrium (UE) principle and indicated that the UE principle should be followed when assigning the traffic demand. The UE principle assumes that: 1) All travelers know the precise cost of each potential route; 2) All travelers will choose the path with the minimum cost. In the TAP, when UE status achieved, all travelers (travel through the same OD pair) would have the same cost no matter which route they choose. What's more, under the UE status, no travelers could reduce their cost by altering their routes. However, in the classical traffic assignment problem (TAP), researchers seldom consider the capacity constraints of links. Thus, in the TAP, some links would be over-saturated when user equilibrium is reached, which is not possible in the real world. Thus, it is important for us to study the link-capacitated traffic assignment (CTAP) problem.

At present, approaches to address the CTAPs can be classified into two types: (a) Lagrangian multipliers and (b) penalty functions \cite{hearn1980bounded,hestenes1969multiplier}. Attempts have been made to improve the augmented Lagrange multiplier (ALM) method and the inner penalty function (IPF) method \cite{GG_CTAPNie}. Readers could refer to \cite{GG_CTAPNie,feng2020efficient} for more information about the CTAP.

The formulation of CTAP is given by the following non-linear programming problem:
\begin{align}
\min\quad & z(x)=\sum_{\forall m \in M} \int_{0}^{Q_m} Cost_m(w) \mathrm{d} w \\
\text { subject to: } & \sum_{k} Q_{k}^{r s}=Demand_{rs} \quad \forall r \in R, s \in S \\
& Q_{k}^{r s} \geq 0 \quad \forall k \in K_{r s}, r \in R, s \in S \\
& Q_m=\sum_{r} \sum_{s} \sum_{k} Q_{k}^{r s} \delta_{m, k}^{r s} \quad \forall m \in M, k \in K_{r s}, r \in R, s \in S \\
& Q_m \leq C_{m} \quad \forall m \in M \label{eq.constrain}
\end{align}
where $Q_m$ denotes the flow on link $m$; $R$, $S\in N$, are the sets of origins and destinations; $Cost_m$ is the cost function of link $m$; $Demand_{rs}$ represents the traffic demand between a certain O-D pair $(r,s)$; $K_{rs}$ denotes the set of paths between a certain O-D pair $(r,s)$; $Q_k^{rs}$ is the flow on path k ($k \in K_{rs}$) between a certain O-D pair $(r,s)$; $\delta_{k,m}^{rs}$ represents the link-path incidence and $C_m$ is the capacity of link $m$. The main difference between the CTAP and TAP is that Eq.(\ref{eq.constrain}) is a necessity in the CTAP.

\section{The proposed algorithms}
\label{sec.3}
\subsection{The capacitated physarum polycephalum inspired algorithm (CPPA)}
\label{sec.3.1}
Recall the basic knowledge of the original PPA mentioned in Section \ref{sec.2.1}. We would find out that no capacity related variables are involved in the original PPA. In other words, we could not control the amount of flow flowing through the edges in the graph. Hence, in this subsection, we would introduce a method (the CPPA) to enable the flow control in the Physarum Solver.   
\subsubsection{Controlling the amount of flow flowing through links}
\label{sec.3.1.1}
Wang \textit{et al.} \cite{GG_PPAMC_ZWang} and Wang \cite{GG_PPAMC_QWang} made preliminary attempts in controlling the amount of flow in the PPA by proposing the concept of basic link flow and the concept of energy flow respectively. The main idea of the two concepts is quite similar. Let's call the flow along different paths from the starting node to the ending node 'stream'. While the PPA iterating, if some links in a path are saturated (which means that the flow along this link reaches its capacity), then the corresponding streams would be separated from the PPA and marked down. Then the capacities would be updated to its original value minus the separated amount of flow. The separated streams would not take part in the next iteration, which weakens the adaptive characteristics of the PPA. What's more, these concepts are quite complex to implement. Also, when there is more than one starting node or ending node, there would be more than one stream flowing through one link. How to separate the corresponding streams would be a problem in this scenario. Thus, controlling the link flow by separating streams that cause saturation from the PPA might not be a good idea. 

In Eq.(\ref{eq.1}), the value of flow $Q_{ij}$ flowing through edge $M_{ij}$ is relevant to conductivity $D_{ij}$, the length $L_{ij}$ of the edge, and the pressure $p_i$ and $p_j$ of the edge's nodes. The value of $L_{ij}$ is mainly dependent on the applications, which means that using $L_{ij}$ for flow control would weaken the robustness of the algorithm. For  $p_i$ and $p_j$, the value of them would change as soon as any flow flowing through node $i$ and $j$ change. Thus, the algorithm might be hard to converge if  $p_i$ and $p_j$ are adopted for flow control. Hence, controlling the link flow using the conductivity $D_{ij}$ would be our choice. Thus, in the CPPA, Eq.(\ref{eq.7}) could be changed to:
\begin{equation}
D_{ij}^{n+1}=\left \{\begin{gathered}
\frac{|Q_{ij}^n|+D_{ij}^n}{2} \ \ for\ Q_{ij}^n \leq C_{ij},  \\
\frac{C_{ij}L_{ij}}{\lvert p_i^n-p_j^n\rvert}\ \ for\ Q_{ij}^n \textgreater C_{ij},  \\
\end{gathered} \right.
\label{eq.Dchanged}
\end{equation}
where $p_i^n$ and $p_j^n$ are the pressure of node $i$ and $j$ at $n^{th}$ iteration; $C_{ij}$ is the capacity of edge $M_{ij}$; $Q_{ij}^n$ is the flow flowing though edge $M_{ij}$ at $n^{th}$ iteration; $D_{ij}^{n+1}$ and $D_{ij}^{n}$ denote the conductivity of the edge $M_{ij}$ at the $n+1$th and the $n$th iteration respectively. 

Eq.(\ref{eq.Dchanged}) is the core of controlling the link flow. On one hand, when the link flow $Q_{ij}$ along edge $M_{ij}$ does not violate the capacity constraint (which means that: $Q_{ij}\leq C_{ij}$), the conductivity $D_{ij}$ still follow the adaptation rule given in Eq.(\ref{eq.7}). On the other hand, when the edge is oversaturated, the conductivity would be re-regulated so that at next iteration the capacity constraint would not be violated. However, when we implement Eq.(\ref{eq.Dchanged}) in our algorithm, we found that the algorithm sometimes did not converge but oscillate in the iterative process. Thus, in order to avoid oscillation and encourage convergence, we introduce a threshold in Eq.(\ref{eq.Dchanged}). Thus, we would have the following adaptation rule of the conductivity in the CPPA: 
\begin{equation}
D_{ij}^{n+1}=\left \{\begin{gathered}
\frac{|Q_{ij}^n|+D_{ij}^n}{2} \ \ for\ Q_{ij}^n \leq k \cdot C_{ij},  \\
\frac{C_{ij}L_{ij}}{\lvert p_i^n-p_j^n\rvert}\ \ for\ Q_{ij}^n \textgreater  k\cdot C_{ij},  \\
\end{gathered} \right.
\label{eq.Dchanged_threshold}
\end{equation}
where $k$ is the pre-defined threshold ($k\in (0,1]$). The influence of the threshold parameter would be discussed in Section \ref{sec.4.1}.

\subsubsection{The procedure of CPPA}
\label{sec.3.1.2}
The procedure of CPPA at $n^{th}$ iteration is written as follows ($L_{ij}$ and $I_0$ are given before the start of iteration):
\begin{enumerate}[Step 1]
	\item Input $D_{ij}^{n}$ from last iteration.
	\item With $L_{ij}$ and $I_0$, solve Eq.(\ref{eq.4}) to obtain $p_i$.
	\item With $p_i$, $L_{ij}$ and $D_{ij}^{n}$, solve Eq.(\ref{eq.1}) to obtain $Q_{ij}$.
	\item Generate $D_{ij}^{n+1}$ for the next iteration according to the equation proposed in the last sub-subsection which is Eq.(\ref{eq.Dchanged_threshold}). 
\end{enumerate}

\subsubsection{A simple example}
\label{sec.3.1.3}
\begin{figure}[htb!]
	\centering
	\includegraphics[width=0.7\textwidth]{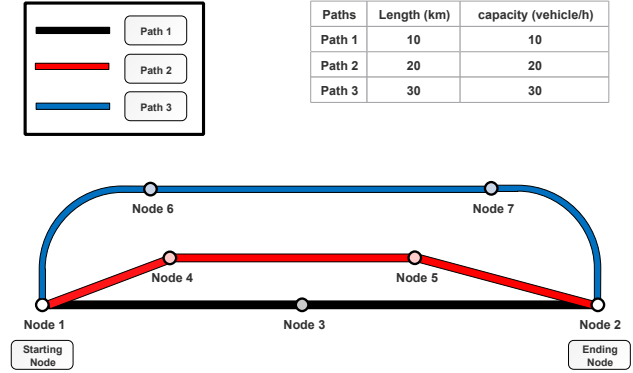}
	\caption{A graph composed of three paths.}
	\label{fig.CPPA-example}
\end{figure}
In this part, both the original PPA and the CPPA would be applied in a small graph (Fig.\ref{fig.CPPA-example}) to show the difference between whether we enable flow controlling in the PPA or not. Fig.\ref{fig.CPPA-example} has three paths with different lengths and capacities. Path 1, which is the path $Node1 \rightarrow Node3 \rightarrow Node2$, is the shortest path of the graph. 

During iterations, we would sequentially adjust the amount of flow which flows in the graph (Fig.\ref{fig.CPPA-example}) through $Node1$ and flows out through $Node2$. We found that the original PPA and the CPPA behaved quite differently. 

\begin{figure}[htb!]
	\centering
	\includegraphics[width=0.8\textwidth]{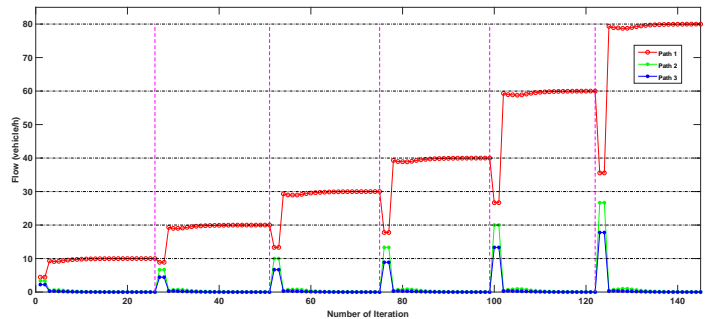}
	\caption{Results of applying the original PPA in a small graph (Fig.\ref{fig.CPPA-example}).}
	\label{fig.CPPAexample-PPA}
\end{figure}

Fig.\ref{fig.CPPAexample-PPA} demonstrates the results of applying the original PPA in the small graph. In this figure, the black horizontal lines are the reference lines and the purple vertical lines mark the moments when we adjust the flow. It can be seen that, in the beginning, when the inflow is 10 $vehicle/h$, the flow in path 1 is 10 $vehicle/h$ while the flow in other paths are $0$ when the original PPA converges. At the $26th$ iteration, which is after the first convergence of the original PPA, we adjust the inflow to 20 $vehicle/h$. However, after several iterations, the flow in path 1 is 20 $vehicle/h$ while the flow in other paths are $0$ when the original PPA converges. The amount of flow in path 1 exceeds its capacity. This phenomenon occurs every time we adjust the inflow, which means that the original PPA does not have the ability to control the amount of flow flowing through certain links as we have mentioned before. Instead, the original PPA could only direct all the inflow to the graph's shortest path which is path 1. 

\begin{table}[htb] 
	\centering
	\caption{Results of applying the CPPA in a small graph (Fig.\ref{fig.CPPA-example}).}
	\resizebox{\textwidth}{12.5mm}{
		\begin{tabular}{ccccc}
			\toprule[0.75pt]   
			Inflow ($vehicle/h$) &Iteration of convergence & \tabincell{c}{The amount of flow \\ on path 1} & \tabincell{c}{The amount of flow \\ on path 2} &\tabincell{c}{The amount of flow \\ on path 3}  \\ \cline{1-5} 
			10&1113&   10 & 0 & 0 \\
			20&3953&   10 & 9.667 & 0.333 \\
			40&6806&    9.439 & 18.378 & 12.183\\
			80&37068 &20.01&29.98&30.01\\
			\bottomrule[0.75pt] 
		\end{tabular}
	}
	\label{table.CPPAexample-CPPA}
\end{table}

When we apply the CPPA in the small graph, link flow could be effectively controlled according to Table \ref{table.CPPAexample-CPPA}.  It can be seen that, in the beginning, when the inflow is 10 $vehicle/h$, the flow in path 1 is 10 $vehicle/h$ while the flow in other paths are $0$ when the CPPA converges. At the $1113th$ iteration, which is after the first convergence of the CPPA, we adjust the inflow to 20 $vehicle/h$. After several iterations, the flow in path 1 is 10 $vehicle/h$ while the flow in paths 2 and 3 are 9.667 and 0.333 $vehicle/h$ respectively when the CPPA converges. The amount of flow in path 1 does not exceed its capacity. At the $3953rd$ iteration, which is after the second convergence of the CPPA, we adjust the inflow to 40 $vehicle/h$. After several iterations, the flow in path 1 is 9.439 $vehicle/h$ while the flow in paths 2 and 3 are 18.378 and 12.183 $vehicle/h$ respectively when the CPPA converges. The amount of flow in path 1 and path 2 do not exceed their capacity. 

This small example shows that the CPPA has the capability to control the link flow. We believe that the accuracy of flow control depends on the convergence criterion of the PPA. In order to achieve a satisfying performance, the CPPA may cost more running time than the original PPA. To be noticed, when the inflow is adjusted to 80 $vehicle/h$, which is higher than the maximum flow of the graph, the CPPA is no longer effective according to Table \ref{table.CPPAexample-CPPA}. Thus, when applying the CPPA, the inflow should not be set to larger than the maximum flow of the graph.

\subsection{The CPPA for the MF problem (CPPA-MF)}
\label{sec.3.2}
In the small example provided in the Section \ref{sec.3.1.3}, we find that: if $1$ to $(k-1)th$ shortest paths of the graph are saturated while the $kth$ shortest path is not, the CPPA would direct the flow remained (which is equal to the inflow minus the total amount of flow in $1$ to $(k-1)th$ shortest paths of the graph) to the $kth$ shortest path of the graph. Based on the above finding and inspired by references \cite{GG_PPAMC_ZWang} and \cite{GG_PPAMC_QWang}, we propose our CPPA-MF:
\begin{enumerate}[1)]
	\item In a given original graph, embed a virtual path connecting the starting node and the ending node (whose length is long enough and whose capacity is large enough)  to create a new graph. Set the inflow of the new graph to a value that is significantly larger than the maximum flow of the original graph but is smaller than the maximum flow of the new graph.
	\item Apply the CPPA in the new graph.
	\item When the CPPA converge, the inflow of the graph minus the amount of flow in the virtual path would be equal to the maximum flow of the original graph. 
\end{enumerate}

\begin{figure}[htb!]
	\centering
	\includegraphics[width=0.8\textwidth]{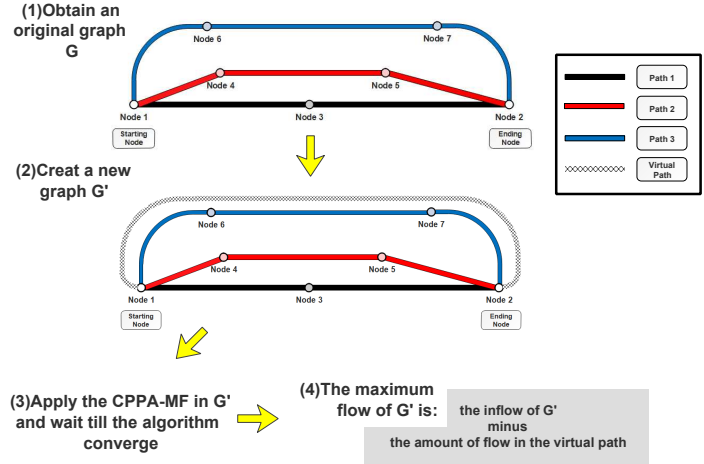}
	\caption{The flow chart of the CPPA-MF.}
	\label{fig.CPPA-MF_flowchart}
\end{figure}

We demonstrate the above procedures in Fig.\ref{fig.CPPA-MF_flowchart}. To be noticed, if in the original graph there exists a link directly connecting the starting and the ending nodes, we need to add a virtual node in the virtual path so that there would not be two paths connecting the starting node and the ending node directly.

We provide the pseudo-code of the CPPA-MF in Algorithm \ref{alg.CPPA-MF}. Normally, we set the length of the virtual path to $100*\sum_{M_{ij}\in M}L_{ij}$ ($M$ is the set of edges in the graph and $L_{ij}$ is the length of edge $M_{ij}$) and the capacity of the virtual path to $100*\sum_{M_{ij}\in M}C_{ij}$ ($C_{ij}$ is the capacity of edge $M_{ij}$). The above settings make sure that the value of the length and the capacity of the virtual path is larger than that of any other path in the original graph. Additionally, we set the amount of inflow to the capacity of the virtual path, which guarantees that the inflow of the new graph would be larger than the maximum flow of the original graph. In the MF problem, we usually have $L_{ij}=1$. The stopping criterion is set to $\sum_{M_{ij}\in M}|(Q_{ij}^{count}-Q_{ij}^{count-1}|<N \cdot \epsilon $ according to \cite{GG_PPAMC_QWang}. The pseudo-code presented here is for solving the MF problem in undirected graphs. If you want to migrate it to the MF problem in directed graphs without two-way edges, replace the equation in step 14) of the algorithm with Eq.(14) in reference \cite{GG_xiaogeTAP} and follow its procedures to regulate the possible illegal value of $Q_{ij}$.

\begin{breakablealgorithm}
	\caption{The proposed CPPA-MF}
	\label{alg.CPPA-MF}
	\begin{algorithmic}[1]
		\begin{small}
			\STATE {//Initialization part}
			\STATE {\textbf{Input:} the statistics of graph $G(N,M)$ and the corresponding set of capacities $C$.}
			\STATE {Embed a virtual path $M_v$ in the graph.}
			\STATE {$D_{ij}\Leftarrow0.5\ (\forall M_{ij}\in M,M_v)$} // Initialize the conductivity of each edge.
			\STATE {$Q_{ij}\Leftarrow0\ (\forall M_{ij}\in M,M_v)$} // Initialize the flow of each edge.
			\STATE {$L_{ij}\Leftarrow 1 (\forall M_{ij}\in M)$} //In the MF problem, we usually have $L_{ij}=1$.
			\STATE {$L_{v}\Leftarrow 100*\sum_{M_{ij}\in M}L_{ij}$} //Set the length of the virtual path.
			\STATE {$C_{v}, I_0\Leftarrow 100*\sum_{M_{ij}\in M}C_{ij}$} //Set the capacity of the virtual path and the inflow of the graph.
			\STATE {$p_i\Leftarrow0\ (\forall i=1,2,\cdots,N)$} // Initialize the pressure at each node.
			\STATE {$k\Leftarrow0.85$}
			\STATE {$count\Leftarrow1$} // Initialize the counting variable.
			\STATE {//Iterative part}  
			\REPEAT 
			\STATE  $p_1\Leftarrow0$ // the pressure of the starting node (Node $1$)
			\STATE  Calculate the pressure of all the nodes in the graph using Eq.(\ref{eq.4}) (N denotes the set of all the nodes in the graph):
			\begin{center}
				\begin{equation*}            
				\sum_{i\in N}{\frac{D_{ij}}{L_{ij}}(p_i-p_j)=\left \{\begin{gathered}
					+I_0\ \ for\ j=1,  \\
					-I_0\ \ for\ j=2,  \\
					0\ \ \ otherwise.  \\
					\end{gathered} \right.}                
				\end{equation*}
			\end{center}
			\STATE  Calculate the flux of all the edges in the graph using Eq.(\ref{eq.1}):
			\begin{center}
				$Q_{ij}\Leftarrow D_{ij}\cdot(p_i-p_j)/L_{ij}.$
			\end{center}
			\STATE  Calculate the conductivity of the next generation using Eq.(\ref{eq.Dchanged_threshold}):
			\begin{center}
				\begin{equation*}
				D_{ij}^{n+1}=\left \{\begin{gathered}
				\frac{|Q_{ij}^{count}|+D_{ij}^{count}}{2} \ \ for\ Q_{ij}^{count} \leq k \cdot C_{ij},  \\
				\frac{C_{ij}L_{ij}}{\lvert p_i-p_j\rvert}\ \ for\ Q_{ij}^{count} \textgreater  k\cdot C_{ij},  \\
				\end{gathered} \right.
				\end{equation*}
			\end{center}
			\STATE  {$count\Leftarrow count+1$} 
			\UNTIL{The given termination criterion is met.}
			\STATE {\textbf{Output:} The maximum flow (MF) of the graph $G$: $MF= I_0-Q_v$.}
		\end{small}
	\end{algorithmic}
\end{breakablealgorithm}

\subsection{The CPPA for the minimum-cost-maximum-flow problem (CPPA-MCMF)}
\label{sec.3.3}
In reference \cite{GG_xiaogeSUP}, Zhang \textit{et al.} proposed a framework using the original PPA to solve system optimization problems. Given the maximum flow of the graph, allocating the flow to certain paths to minimize the total cost of the system would be a system optimization problem. The minimum total cost mentioned above would be the minimum cost under the maximum flow of the graph, which is also the solution to the MCMF problem. Thus, we consider using the CPPA-MF to calculate the maximum flow of the graph first and then using the CPPA to calculate the minimum cost under the maximum flow of the graph. 

As is pointed out in \cite{GG_xiaogeSUP}, in order to solve system optimization problems, we need to replace the $L_{ij}$ in the original PPA to the following equation:
\begin{equation}
L_{ij}=UniCost(Q_{ij})+Q_{ij}\cdot\frac{\mathrm{d}UniCost(Q_{ij})}{\mathrm{d}Q_{ij}},
\label{eq.SUP_length}
\end{equation}
where $UniCost(Q_{ij})$ is the cost function for per unit of flow on edge $M_{ij}$. In order to simplify the problem, we consider the cost function for per unit of flow on edge $M_{ij}$ in the MCMF problem is a constant $Cost_{ij}$. Thus, we have:
\begin{equation}
L_{ij}=Cost_{ij}.
\label{eq.SUP_cost}
\end{equation}

Therefore, we propose the CPPA-MCMF algorithm:
\begin{enumerate}[1)]
	\item Given a graph, apply the CPPA-MF first to calculate the maximum flow (MF) of the graph.
	\item Apply the CPPA with Eq.(\ref{eq.SUP_cost}) in the graph.
	\item When the CPPA converge, the minimum cost (MC) under the the maximum flow (MF) of the graph would be $MC=\sum_{M_{ij} \in M}Q_{ij}\cdot Cost_{ij}$.
	\item Output the solutions of the MCMF problem: MF, MC and the flow allocation $Q_{ij} (\forall M_{ij} \in M)$. 
\end{enumerate}

We also provide the pseudo-code of the CPPA-MCMF in Algorithm \ref{alg.CPPA-MCMF}. The termination criterion is the same as Algorithm \ref{alg.CPPA-MF}. 
\begin{breakablealgorithm}
	\caption{The proposed CPPA-MCMF}
	\label{alg.CPPA-MCMF}
	\begin{algorithmic}[1]
		\begin{small}
			\STATE {\textbf{Input:} the statistics of graph $G(N,M)$ and the corresponding set of capacities $C$ and cost function for per unit of flow $Cost$.}
			\STATE {Execute the CPPA-MF (Algorithm \ref{alg.CPPA-MF}) and obtain the maximum flow (MF).}
			\STATE {//Initialization part}
			\STATE {$D_{ij}\Leftarrow0.5\ (\forall M_{ij}\in M)$} // Initialize the conductivity of each edge.
			\STATE {$Q_{ij}\Leftarrow0\ (\forall M_{ij}\in M)$} // Initialize the flow of each edge.
			\STATE {$L_{ij}\Leftarrow Cost_{ij} (\forall M_{ij}\in M)$} //Set $L_{ij}$ according to Eq.(\ref{eq.SUP_cost}).
			\STATE {$p_i\Leftarrow0\ (\forall i=1,2,\cdots,N)$} // Initialize the pressure at each node.
			\STATE {$k\Leftarrow0.85$}
			\STATE {$count\Leftarrow1$} // Initialize the counting variable.
			\STATE {//Iterative part}  
			\REPEAT 
			\STATE  $p_1\Leftarrow0$ // the pressure of the starting node (Node $1$)
			\STATE  Calculate the pressure of all the nodes in the graph using Eq.(\ref{eq.4}) (N denotes the set of all the nodes in the graph):
			\begin{center}
				\begin{equation*}            
				\sum_{i\in N}{\frac{D_{ij}}{L_{ij}}(p_i-p_j)=\left \{\begin{gathered}
					+MF\ \ for\ j=1,  \\
					-MF\ \ for\ j=2,  \\
					0\ \ \ otherwise.  \\
					\end{gathered} \right.}                
				\end{equation*}
			\end{center}
			\STATE  Calculate the flux of all the edges in the graph using Eq.(\ref{eq.1}):
			\begin{center}
				$Q_{ij}\Leftarrow D_{ij}\cdot(p_i-p_j)/L_{ij}.$
			\end{center}
			\STATE  Calculate the conductivity of the next generation using Eq.(\ref{eq.Dchanged_threshold}):
			\begin{center}
				\begin{equation*}
				D_{ij}^{n+1}=\left \{\begin{gathered}
				\frac{|Q_{ij}^{count}|+D_{ij}^{count}}{2} \ \ for\ Q_{ij}^{count} \leq k \cdot C_{ij},  \\
				\frac{C_{ij}L_{ij}}{\lvert p_i-p_j\rvert}\ \ for\ Q_{ij}^{count} \textgreater  k\cdot C_{ij},  \\
				\end{gathered} \right.
				\end{equation*}
			\end{center}
			\STATE  {//If the cost function for per unit of flow on edge $M_{ij}$ in the MCMF problem is not considered as a constant, use Eq.(\ref{eq.SUP_length}) to update $L_{ij}$ here.} 
			\STATE  {$count\Leftarrow count+1$} 
			\UNTIL{The given termination criterion is met.}
			\STATE {The minimum cost (MC) under the the maximum flow of the graph would be: $MC=\sum_{M_{ij} \in M}Q_{ij}\cdot Cost_{ij}$.}
			\STATE {\textbf{Output:} $MC, MF, Q_{ij} (\forall M_{ij} \in M)$.}
		\end{small}
	\end{algorithmic}
\end{breakablealgorithm}

\subsection{The CPPA for the link-capacitated traffic assignment problem (CPPA-CTAP)}
\label{sec.3.4}
Zhang \textit{et al.} \cite{GG_xiaogeTAP} firstly applied the original PPA to solve the TAP. However, it could not be applied to a traffic network with two-way characteristics and also could not distinguish flux in different O-D pairs \cite{GG_PPATAP}. Xu \textit{et al.} improved Zhang's method and proposed the framework for TAP based on the original PPA \cite{GG_PPATAP}. Though both Zhang's and Xu \textit{et al.}'s methods could solve the TAP successfully, the solutions gave by their algorithms still contain over-saturated links, which means that their methods could not be applied to the CTAP problem. 

Thus, we proposed the first PPA-based algorithm for CTAP, the CPPA-CTAP, by replacing the original PPA with the CPPA in Xu \textit{et al.}'s method \cite{GG_PPATAP}. The pseudo-code of the CPPA-CTAP is provided in Algorithm \ref{alg.CPPA-CTAP}. The only difference between the CPPA-CTAP and Xu \textit{et al.}'s method \cite{GG_PPATAP} is that: Xu \textit{et al.} used Eq.(\ref{eq.7}) instead of Eq.(\ref{eq.Dchanged_threshold}) to calculate the conductivity in step 15) in Algorithm \ref{alg.CPPA-CTAP}. Due to space limitations, we omit a detailed explanation of this algorithm. Readers could refer to \cite{GG_PPATAP} for more information.

\begin{breakablealgorithm}
	\caption{CPPA-CTAP}
	\label{alg.CPPA-CTAP}
	\begin{algorithmic}[1]
		\begin{small}
			\STATE {//$R$ is the set of orientation nodes $r$. $Q^r$ is the flow matrix of the subnetwork oriented from node $r$. $L^0$ is the free flow travel time matrix. $C^0$ is the link capacity matrix.}
			\STATE {//Initialization part}
			\STATE {$D_{ij}^r\Leftarrow0.5\ (\forall i,j=1,2,\cdots,N \land C_{ij}^0\leq 0, \forall r\in R)$}
			\STATE {$D_{ij}^r\Leftarrow0\ (\forall i,j | C_{ij}^0 = inf)$} 
			\STATE {$Q_{ij}^r\Leftarrow0\ (\forall i,j=1,2,\cdots,N, \forall r\in R)$} 
			\STATE {$Q_{ij}^{all_1}\Leftarrow0\ (\forall i,j=1,2,\cdots,N$ //the sum of all the subnetwork flow matrices} 
			\STATE {$L_{ij}^1\Leftarrow L_{ij}^0 (\forall i=1,2,\cdots,N)$}
			\STATE {$count\Leftarrow1$}
			\STATE {//Iterative part}  
			\REPEAT 
			\FOR {$r\in R$}
			\STATE  $p_r\Leftarrow0$ // the pressure of the starting node (Node $r$)
			\STATE  Calculate the pressure of all the nodes in the graph ($V$ denotes the set of all the edges in the graph, $S_{rs}$ is the set of the OD pairs $(r,s)$, $I_{rs}$ represents the travel demand of the OD pairs $(r,s)$, $I_{r}$ is the total amount of flow orients from node $r$).:
			\begin{center}
				\begin{equation*}            
				\sum_{i\in V}{\left(\frac{D_{ij}^r}{L_{ij}}+\frac{D_{ji}^r}{L_{ji}}\right)(p_i-p_j)=\left \{\begin{gathered}
					+I_r\ \ for\ j=r,  \\
					-I_{rs}\ \ \forall s\in S_{rs},  \\
					0\ \ \ otherwise.  \\
					\end{gathered} \right.}                
				\end{equation*}
			\end{center}
			\STATE  Calculate the flux of every edge:
			\begin{center}
				\begin{equation*}            
				Q_{ij}^r=\left \{\begin{gathered}
				\frac{D_{ij}^r}{L_{ij}}(p_i-p_j), \ \ \frac{D_{ij}^r}{L_{ij}}(p_i-p_j)>0  \\
				0,\ \ \ otherwise.  \\
				\end{gathered} \right.            
				\end{equation*}
			\end{center}
			\STATE  Calculate the conductivity of the next generation using Eq.(\ref{eq.Dchanged_threshold}):
			\begin{center}
				\begin{equation*}
				D_{ij}^{n+1}=\left \{\begin{gathered}
				\frac{|Q_{ij}^{count}|+D_{ij}^{count}}{2} \ \ for\ Q_{ij}^{count} \leq k \cdot \frac{Q_{ij}^{count}}{Q_{all}^{count}}C_{ij},  \\
				\frac{C_{ij}L_{ij}}{\lvert p_i-p_j\rvert}\ \ for\ Q_{ij}^{count} \textgreater  k\cdot \frac{Q_{ij}^{count}}{Q_{all}^{count}}C_{ij},  \\
				\end{gathered} \right.
				\end{equation*}
			\end{center}
			\ENDFOR
			\STATE     {$Q_{ij}^{all_{count+1}}\Leftarrow \sum_{r\in R} Q_{ij}^r, \forall i,j \in N$.}
			\STATE Update  $L_{ij}$ as follows :    
			\begin{equation}
			\label{eq.OPPA-TAP-L}
			L_{ij}^{count+1}= \frac{L_{ij}^{count}+t_a(Q_{ij}^{all_{count+1}})}{2}
			\end{equation}
			\STATE  {$count\Leftarrow count+1$.}
			\UNTIL{The algorithm converges.}
		\end{small}
	\end{algorithmic}
\end{breakablealgorithm}  
In this paper, the $RGAP$ \cite{GG_xiaogeTAP} is used to measure the convergence:
\begin{equation}
RGAP=1-\frac{\sum_{\forall(r,s)\in S_{rs}}Demand_{rs}\cdot p_{min}^{rs}}{\sum_{i}\sum_{j}Q_{ij}\cdot t_{ij}},
\end{equation}
where $S_{rs}$ is the set of OD pairs; $Demand_{rs}$ represents the travel demand of a certain OD pair $(r,s)$; $p_{min}^{rs}$ is the total travel time of the shortest path between a certain OD pair $(r,s)$; $Q_{ij}$ and $t_{ij}$ denotes the flow and travel time along the edge $M_{ij}$ when the algorithm terminates \cite{GG_xiaogeTAP}. 

\section{Experiments}
\label{sec.4}
In the simple example (Section \ref{sec.3.1.3}), the effectiveness of the CPPA has been preliminarily tested. In this part, we aim to test the effectiveness of the proposed algorithms, i.e., the CPPA-MF, the CPPA-MCMF, and the CPPA-CTAP algorithm. All the algorithms are tested through computer simulations using Matlab R2017b on an Intel Core i5-7300 CPU (2.5GHz) with 8 GB RAM under Windows 10.
\subsection{Testing the CPPA-MF}
\label{sec.4.1}
In this part, we firstly test the proposed CPPA-MF in undirected graphs, and then in directed graphs. In Section \ref{sec.4.1.3}, we test the influence of the threshold parameter $k$ on the algorithm.
\subsubsection{Undirected graph}
\label{sec.4.1.1}
\textbf{Data Set 1:} According to the recommendation in \cite{GG_PPAMC_ZWang}, we test our algorithm in the 'elist' instances. These instances containing grids-on-pipe graphs were generated by G. Skorobohatyi using the program RMFGEN \cite{GG_PPAMC_ZWang}, which could be found in \cite{dataset-MF}. Some basic information of the dataset is provided in Table \ref{table.datasetMF-un}.

\begin{table}[htb] 
	\centering
	\caption{Basic information of the 'elist' instances}
	\resizebox{0.8\textwidth}{15mm}{
		\begin{tabular}{cccc}
			\toprule[0.75pt]   
			Instances &Number of nodes & Number of edges & (Starting|Ending) nodes \\ \cline{1-4} 
			elist96&96&187&(1|96)\\
			elist160&160&285&(1|160)\\
			elist200&200&483&(1|200)\\
			elist500&500&1040&(1|500)\\
			elist640&640&3037&(1|640)\\
			\bottomrule[0.75pt] 
		\end{tabular}
	}
	\label{table.datasetMF-un}
\end{table}

\textbf{Baseline Algorithms:} The Wang's methods \cite{GG_PPAMC_ZWang}, the linear programming (LP) method--"simplex method" \cite{GG_PPAMC_ZWang}, the Ford-Fulkerson (FF)  algorithm are the baseline algorithms for our test. The stopping criterion $\epsilon$ for the CPPA-MF is set to $5e-5$.

\begin{table}[htb] 
	\centering
	\caption{Results of undirected graph testing}
	\resizebox{\textwidth}{17.5mm}{
		\begin{tabular}{ccccccc}
			\toprule[0.75pt]
			\multirow{2}{*}{Instances}&CPPA-MF(Ours)&
			\multicolumn{2}{c}{PS (Wang's \cite{GG_PPAMC_ZWang})}& AA(Wang's \cite{GG_PPAMC_ZWang}) &LP&FF\\
			\cline{3-4}
			&Time(second)&Gap&Time(second)&Time(second)&Time(second)&Time(second)\\ \hline
			elist96&	4.0E+00&	8.2E+00$^*$&	4.9E+01$^*$&	3.5E+01$^*$&	2.1E+00$^*$&	3.4E-03	\\
			elist160&	5.8E+00&	9.5E+00$^*$&	7.0E+01$^*$&	4.0E+01$^*$&	4.4E+00$^*$&	3.1E-03	\\
			elist200&	9.3E-01&	1.2E+01$^*$&	1.0E+02$^*$&	7.5E+01$^*$&	5.2E+00$^*$&	2.8E-03	\\
			elist500&	1.9E+01&	1.1E+01$^*$&	2.0E+02$^*$&	8.1E+01$^*$&	6.6E+00$^*$&	3.0E-03	\\
			elist640&	7.8E-01&	1.2E+01$^*$&	3.3E+02$^*$&	1.4E+02$^*$&	8.1E+00$^*$&	5.7E-03	\\\bottomrule[0.75pt] 
		\end{tabular}
	}
	\footnotesize{1)$^*$ The statistics of the evaluation metrics of the model quote from paper \cite{GG_PPAMC_ZWang}.} \footnotesize{2)'Gap' denotes $|MF_{standard}-MF_{algorithm}|$.}
	\label{table.resultMF-un}
\end{table}

We will use the execution time to evaluate the convergence speed of different algorithms. The results presented in Table \ref{table.resultMF-un} is the average value of 10 times running. As shown in Table.\ref{table.resultMF-un}, the execution time for different algorithms in solving the MF problem is provided. Overall, the FF algorithm is the fastest and the PS algorithm is the slowest. Being the slowest algorithm, the accuracy of the PS algorithm is also not satisfying, as it has quite a large gap when converges. 

The proposed CPPA-MF achieves good performance in this experiment. In 'elist200' and 'elist640', the CPPA-MF only take 0.93 and 0.78 seconds to converge, which are approximately 82\% and 90\% faster than that of the LP algorithm, taking the second place. In 'elist96', 'elist160', and 'elist640', the CPPA-MF rank the third. In 'elist96' to 'elist640', the CPPA-MF is approximately 91.9\%, 91.7\%, 99.1\%, 90.4\%, 99.8\% faster than that of the PS algorithm (a PPA-based algorithm), respectively; it is also 76.8\%, 76.3\%, 84.7\%, 86.3\%, 91.3\% faster than that of the AA algorithm (an improved PPA-based algorithm), respectively. In conclusion, though the CPPA-MF is slower than the classic MF algorithm (the FF algorithm), it is faster than both of the PPA-based MF algorithms. In some instances, it could even beat the famous linear programming (LP) method. 

The great performance of the CPPA-MF indicates that adopting the CPPA framework in the MF problem is valid. The success of the CPPA-MF also suggests that the CPPA could control link flow in the PPA. As the CPPA is faster than both the PPA-based algorithms, the CPPA framework is superior to the framework proposed by Wang \textit{et al.} \cite{GG_PPAMC_ZWang} in terms of effectiveness and efficiency. The reason why the CPPA-MF is slower than the classic algorithm is that the time complexity of both the original PPA and the CPPA is $O(n^3)$ as they need to solve the linear Eq.(\ref{eq.4}). It will be our future work to reduce the time complexity of the original PPA and the CPPA.

\subsubsection{Directed graph}
\label{sec.4.1.2}
\textbf{Data Set 2:} The graphs used to evaluate the proposed algorithm are randomly generated directed graph without rings and negative weights edges. For any two nodes in the graph, they have 0.7 probability to be connected by a one-way link and have 0.3 probability not to be connected. The graph size ($N$) ranges from 100 to 800. For each graph size, 10 graphs are randomly generated with the edge's weights equalling to 1 and edge's capacity ranging from 1 to 10 to simulate different topologies. The value presented in Table \ref{table.resultMF-dir} is the average value of the 10 cases. 

\textbf{Baseline Algorithms:} The Wang's method \cite{GG_PPAMC_QWang} and the Ford-Fulkerson (FF) algorithm are the baseline algorithms for our test. The stopping criterion $\epsilon$ for the CPPA-MF is set to $5e-5$ when the graph size is smaller than 400 and is set to $1e-4$ when the graph size is larger than 400.

\begin{table}[htb] 
	\centering
	\caption{Results of directed graph testing.}
	\resizebox{0.8\textwidth}{24mm}{
		\begin{tabular}{cccc}
			\toprule[0.75pt]
			\multirow{2}{*}{Graph size}&CPPA-MF(Ours)&Wang's method \cite{GG_PPAMC_QWang} & FF\\
			&Time(second)&Time(second)&Time(second)\\ \hline
			100&	0.06&	0.02&	0.01 	\\
			200&	0.43&	0.11&	0.03 	\\
			300&	1.02&	0.20&	0.11 	\\
			400&	3.25&	0.82&	0.27 	\\
			500&	3.89&	1.20&	0.52 	\\
			600&	5.34&	1.68&	0.86 	\\
			700&	7.16&	2.59&	1.33 	\\
			800&	8.20&	3.25&	1.97 	\\
			\bottomrule[0.75pt] 
		\end{tabular}
	}
	\label{table.resultMF-dir}
\end{table}
Table \ref{table.resultMF-dir} provides the results of directed graph testing. Overall, the FF algorithm is the fastest and the CPPA-MF algorithm is the slowest. Though CPPA-MF is slower than Wang's method, the Wang's method could not be applied to undirected graphs, which means that our algorithm is more robust than Wang's method at a price of low efficiency. On average, the speed gap between the CPPA-MF and Wang's method \cite{GG_PPAMC_QWang} is within an order of magnitude, which is acceptable when the graph size is not too large. It will be our future work to improve the efficiency of the CPPA-MF algorithm.

\subsubsection{The influence of threshold parameter $k$}
\label{sec.4.1.3}

\begin{figure}[htb!]
	\centering
	\includegraphics[width=0.9\textwidth]{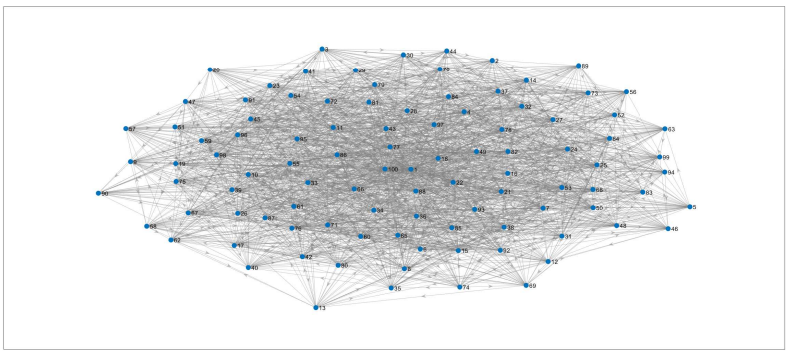}
	\caption{N100 instance.}
	\label{fig.dataset_Node100}
\end{figure}

\textbf{Data Set 3:} We randomly select three graphs with 100, 300 and 600 nodes generated in dataset 2 as our dataset. Table \ref{table.datasetMF_parameters} provides some basic information of the graphs and Fig.\ref{fig.dataset_Node100} demonstrates the 'Node100' instance. The value presented in  Table \ref{table.resultMF_parameter} is the average value of 10 times running. 
\begin{table}[htb] 
	\centering
	\caption{Basic information of dataset 3}
	\begin{tabular}{cccc}
		\toprule[0.75pt]   
		Instances &Number of nodes & Number of edges & (Starting|Ending) nodes \\ \cline{1-4} 
		N100&100&2085&(1|100)\\
		N300&300&18320&(1|300)\\
		N600&600&72806&(1|600)\\
		\bottomrule[0.75pt] 
	\end{tabular}
	\label{table.datasetMF_parameters}
\end{table}

\begin{table}[htb] 
	\centering
	\caption{Running time of the CPPA-MF algorithm with different $k$ in dataset 3.}
	\resizebox{\textwidth}{10mm}{
		\begin{tabular}{ccccccccccc}
			\toprule[0.75pt]
			\multirow{3}{*}{Instances}&\multicolumn{10}{c}{$k$}\\
			&	0.1&	0.2&	0.3&	0.4&	0.5&	0.6&	0.7&	0.8&	0.9&	1	\\\cline{2-11}
			&	Time(second)&	Time(second)&	Time(second)&	Time(second)&	Time(second)&	Time(second)&	Time(second)&	Time(second)&	Time(second)&	Time(second)	\\\hline
			N100&	\XSolidBrush&	\XSolidBrush&	\XSolidBrush&	\XSolidBrush&	5.96E-02&	5.78E-02&	5.99E-02&	6.24E-02&	6.01E-02&	-	\\
			N300&	\XSolidBrush&	1.31E+00&	1.29E+00&	1.33E+00&	1.34E+00&	4.55E+00&	1.30E+00&	1.31E+00&	1.31E+00&	-	\\
			N600&	6.76E+00&	6.83E+00&	6.89E+00&	3.51E+00&	7.00E+00&	8.47E+00&	6.27E+00&	5.73E+00&	5.60E+00&	-	\\
			\bottomrule[0.75pt] 
		\end{tabular}
	}
	\scriptsize{NOTE: $-$ denotes that the algorithm did not coverage after a longer time running; \XSolidBrush\quad represents that the algorithm converged to a wrong MF.}
	\label{table.resultMF_parameter}
\end{table}

Table \ref{table.resultMF_parameter} provides the running time of the CPPA-MF algorithm with different $k$. According to Table \ref{table.resultMF_parameter}, the CPPA-MF could not converge when $k=1$ (which means no threshold in CPPA) in all of the three instances, which indicates that the introduction of the threshold $k$ is very necessary. In 'N100', when $k=0.1,0.2,0.3,$ and $0.4$, the CPPA-MF could not generate the MF accurately, which also happened in 'N300' when $k=0.1$. However, when the algorithm could generate a correct solution, the running-time gap between the CPPA-MF when different $k$ adopted is within an order of magnitude. Thus, the accuracy of the CPPA-MF algorithm is sensitive to the value of the threshold $k$ but the running time of the CPPA-MF is not. When $k$ is too small, the CPPA-MF may not be able to generate a correct solution; however, when $k$ is too large, the CPPA-MF may not be able to converge in a certain amount of running time. Thus, according to our experience, we suggest that $k\in [0.7,1)$. When applied in a problem needing high accuracy, $k$ should be set to a number as close to 1 as possible while maintaining the fast convergence of the algorithm.

\subsection{Testing the CPPA-MCMF}
\label{sec.4.2}
\begin{table}[htb] 
	\centering
	\caption{Basic information of dataset 4}
	\resizebox{\textwidth}{19mm}{
		\begin{tabular}{cccccc}
			\toprule[0.75pt]   
			Instances &Number of nodes & Number of edges & (Starting|Ending) nodes &\tabincell{c}{Maximum flow \\ with minimum cost}  &Minimum cost\\ \hline
			M100&100&2085&(1|100)&297&670\\
			M200&200&8122&(1|200)&570&1289\\
			M300&300&18320&(1|300)&908&2052\\
			M400&400&32290&(1|400)&1169&2644\\
			M500&500&50555&(1|500)&1482&3359\\
			M600&600&72806&(1|600)&1723&3912\\
			M700&700&98704&(1|700)&2076&4684\\
			M800&800&128718&(1|800)&2308&5221\\
			\bottomrule[0.75pt] 
		\end{tabular}
	}
	\label{table.datasetMCMF}
\end{table}
\textbf{Data Set 4:} The graphs used to evaluate the proposed algorithm are randomly generated directed graph without rings and negative weights edges. For any two nodes in the graph, they have 0.7 probability to be connected by a one-way link and have 0.3 probability to not be connected. The graph size ($N$) ranges from 100 to 800. For each graph size, one graph is randomly generated with the edge's unit-cost ranging from 1 to 10 and edge's capacity ranging from 1 to 10 to simulate different topologies. Table \ref{table.datasetMCMF} provides some basic information of the graphs. The value presented in Table \ref{table.resultMF-dir} is the average value of the 10 times running.

\textbf{Baseline Algorithms:} The Bellman-Ford (BF) algorithm \cite{BF_algorithm} and an open-source MCMF Matlab algorithm \cite{zhang2017finding} (We will call it W Zhang's method below.) are the baseline algorithms for our test.

\begin{table}[htb] 
	\centering
	\caption{Results of testing the CPPA-MCMF.}
	\resizebox{\textwidth}{20mm}{
		\begin{tabular}{ccccccc}
			\toprule[0.75pt]
			\multirow{2}{*}{Instances}&\multicolumn{4}{c}{CPPA-MCMF(Ours)}&BF&W Zhang's method\\\cline{2-5}
			&Time(second)&\tabincell{c}{Corresponding \\ termination criterion $\epsilon$}&Minimum cost&\tabincell{c}{Maximum flow \\ with minimum cost}&Time(second)&Time(second)\\\hline
			M100&	0.06&	3.00E-03&	670.29&	297.00&	0.12&	0.10 	\\
			M200&	0.67&	3.00E-04&	1289.56&	570.00&	0.70&	0.60 	\\
			M300&	1.73&	1.00E-04&	2052.52&	908.00&	2.13&	1.99 	\\
			M400&	3.78&	1.00E-04&	2644.57&	1169.00&	4.94&	5.38 	\\
			M500&	5.62&	8.00E-05&	3359.37&	1482.00&	10.04&	11.76 	\\
			M600&	9.18&	7.00E-05&	3912.35&	1723.00&	20.40&	20.72 	\\
			M700&	15.07&	6.00E-05&	4684.25&	2076.00&	30.96&	34.04 	\\
			M800&	17.81&	1.00E-04&	5221.61&	2308.00&	48.12&	50.39 	\\
			\bottomrule[0.75pt] 
		\end{tabular}
	}
	\label{table.resultMCMF}
\end{table}
Table \ref{table.resultMCMF} provides the results of testing the CPPA-MCMF.  According to Tables and \ref{table.datasetMCMF} and \ref{table.resultMCMF}, the CPPA-MCMF successfully calculated the maximum flow of all the instances. As for the minimum cost under the maximum flow, the result generated by CPPA-MCMF is the same as the correct answer, with a calculation error smaller than 1. The above findings indicate that applying the CPPA in MCMF problem solving is effective, which further demonstrates the validation of CPPA in link flow controlling.

Table \ref{table.datasetMCMF} also shows that the CPPA-MCMF is the fastest algorithm among the three. What's more, the increase in running time of the CPPA-MF is significantly lower than the increase in running time of the other two algorithms as the scale of the graph increases. Thus, the CPPA-MCMF is superior to the baseline algorithms in terms of efficiency.

\begin{table}[htb] 
	\centering
	\caption{Results of testing the CPPA-MCMF's sensitivity to the stopping criterion $\epsilon$ in 'M100' instance.}
	\resizebox{\textwidth}{17.5mm}{
		\begin{tabular}{cccccc}
			\toprule[0.75pt]
			\multirow{2}{*}{Termination criterion $\epsilon$} & \multirow{2}{*}{Time(second)} & \multicolumn{2}{c}{Correct answers} & \multicolumn{2}{c}{Ours} \\\cline{3-6}
			&  & \tabincell{c}{Maximum flow \\ with minimum cost} & Minimum cost & \tabincell{c}{Maximum flow \\ with minimum cost} & Minimum cost \\\hline
			5.00E-03 & 0.03 & \multirow{6}{*}{297} & \multirow{6}{*}{670} & \textbf{297} & \textbf{670.58} \\
			1.00E-03 & 0.08 &  &  & \textbf{297} & \textbf{670.26} \\
			5.00E-04 & 0.08 &  &  & \textbf{297} & \textbf{670.25} \\
			1.00E-04 & 0.18 &  &  & \textbf{297} &\textbf{670.11}\\
			5.00E-05 & 0.39 &  &  & \textbf{297} & \textbf{670.07} \\
			1.00E-06 & 0.83 &  &  & \textbf{297} & \textbf{670.00} \\\bottomrule[0.75pt] 
		\end{tabular}
	}
	\label{table.resultMCMF_M100}
\end{table}
\begin{table}[htb] 
	\centering
	\caption{Results of testing the CPPA-MCMF's sensitivity to the stopping criterion $\epsilon$ in 'M300' instance.}
	\resizebox{\textwidth}{17.5mm}{
		\begin{tabular}{cccccc}
			\toprule[0.75pt]
			\multirow{2}{*}{Termination criterion $\epsilon$} & \multirow{2}{*}{Time(second)} & \multicolumn{2}{c}{Correct answers} & \multicolumn{2}{c}{Ours} \\\cline{3-6}
			&  & \tabincell{c}{Maximum flow \\ with Minimum cost} & minimum cost & \tabincell{c}{Maximum flow \\ with minimum cost} & Minimum cost \\\hline
			5.00E-03 & 0.24 & \multirow{6}{*}{908} & \multirow{6}{*}{2052} & 914 & 2072.94 \\
			1.00E-03 & 0.35 &  &  & 909 & 2058.98 \\
			5.00E-04 & 0.49 &  &  & \textbf{908} & 2056.26 \\
			1.00E-04 & 1.73 &  &  & \textbf{908} & \textbf{2052.52 }\\
			5.00E-05 & 2.08 &  &  & \textbf{908} & \textbf{2052.25} \\
			1.00E-06 & 3.42 &  &  & \textbf{908} & \textbf{2052.01} \\\bottomrule[0.75pt] 
		\end{tabular}
	}
	\label{table.resultMCMF_M300}
\end{table}
\begin{table}[htb] 
	\centering
	\caption{Results of testing the CPPA-MCMF's sensitivity to the stopping criterion $\epsilon$ in 'M600' instance.}
	\resizebox{\textwidth}{17.5mm}{
		\begin{tabular}{cccccc}
			\toprule[0.75pt]
			\multirow{2}{*}{Termination criterion $\epsilon$} & \multirow{2}{*}{Time(second)} & \multicolumn{2}{c}{Correct answers} & \multicolumn{2}{c}{Ours} \\\cline{3-6}
			&  & \tabincell{c}{Maximum flow \\ with Minimum cost} & minimum cost & \tabincell{c}{Maximum flow \\ with minimum cost} & Minimum cost \\\hline
			5.00E-03 & 1.90 & \multirow{6}{*}{1723} & \multirow{6}{*}{3912} & 1743 & 3978.64 \\
			1.00E-03 & 2.95 &  &  & 1734 & 3952.41 \\
			5.00E-04 & 10.73 &  &  & 1728 & 3928.01 \\
			1.00E-04 & 10.58 &  &  & 1724 & 3915.13 \\
			5.00E-05 & 9.10 &  &  & \textbf{1723} & \textbf{3912.25} \\
			1.00E-06 & 18.97 &  &  & \textbf{1723} & \textbf{3912.00}\\
			\bottomrule[0.75pt] 
		\end{tabular}
	}
	\label{table.resultMCMF_M600}
\end{table}
In our experiments, we found that the running time of the CPPA-MCMF also depends on the termination criterion. Thus, we test the CPPA-MCMF algorithm's sensitivity to the stopping criterion $\epsilon$ here. The results are presented in Tables \ref{table.resultMCMF_M100}, \ref{table.resultMCMF_M300}, and \ref{table.resultMCMF_M600}. As shown in the tables, if $\epsilon$ is large, the CPPA-MCMF would not be able to generate the correct answers. For example, according to Table \ref{table.resultMCMF_M600}, in the 'M600' instance, when $\epsilon$ is beyond $5e-5$, the MF and the MC generated by the CPPA-MCMF are not correct. On the other hand, if $\epsilon$ is too small, the running time of CPPA-MCMF would increase significantly. Thus, given a certain error tolerance, how to choose the proper $\epsilon$ to maintain the CPPA-MCMF's accuracy while achieving fast convergence would be our future study.

\subsection{Testing the CPPA-CTAP}
\label{sec.4.3}
\begin{figure}[htb!]
	\centering
	\includegraphics[width=0.6\textwidth]{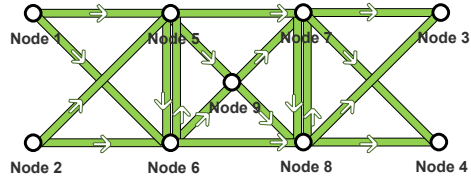}
	\caption{The Hearn graph.}
	\label{fig.Hearn}
\end{figure}
\begin{table}[htb!] 
	\centering
	\caption{Results of testing the CPPA-CTAP in Hearn graph.}
	\resizebox{0.7\textwidth}{40mm}{
		\begin{tabular}{ccccc}
			\toprule[0.75pt]
			Link & \tabincell{c}{(Initial|Terminal)\\ Nodes} & Capacity & Flow(Xu \textit{et al.}'s method \cite{GG_PPATAP}) & Flow(Ours) \\\hline
			1 & (1,5) & 12.02 & \textbf{12.33465} & 12.02 \\
			2 & (1,6) & 18.02 & 17.66535 & 16.55143 \\
			3 & (2,5) & 43.59 & \textbf{69.99997} & 43.28814 \\
			4 & (2,6) & 26.59 & 3.17E-05 & 22.42615 \\
			5 & (5,6) & 50 & 10.60365 & 10.59143 \\
			6 & (5,7) & 25 & 23.55885 & 4.985856 \\
			7 & (5,9) & 35 & \textbf{48.17213 }& 34.01657 \\
			8 & (6,5) & 50 & 0 & 0 \\
			9 & (6,8) & 25 & 3.453652 & 23.06821 \\
			10 & (6,9) & 35 & 24.81538 & 20.78748 \\
			11 & (7,3) & \textbf{25} & \textbf{40} & 0 \\
			12 & (7,4) & 24 & 23.628 & \textbf{24.53008} \\
			13 & (7,8) & 50 & 2.81E-21 & 8.24E-11 \\
			14 & (8,2) & 39 & 3.24E-07 & 0 \\
			15 & (8,4) & 43 & 36.372 & 41.1842 \\
			16 & (8,7) & 50 & 0 & 1.428571 \\
			17 & (9,7) & 35 & \textbf{40.06916} & 23.82994 \\
			18 & (9,8) & 25 & \textbf{32.91835} & \textbf{25.25885}\\\bottomrule[0.75pt] 
		\end{tabular}}
	\\\footnotesize{Note: The bold data denotes the over-saturated flow.}
	\label{table.resultCTAP}
\end{table}
In this part, we test the proposed CPPA-CTAP. Firstly, we applied both the original-PPA based UE algorithm \cite{GG_PPATAP} (we will call it Xu et al's method below.) and the CPPA-CTAP algorithm in the Hearn graph \cite{hearn1980bounded}. The structure of the Hearn graph is presented in Fig.\ref{fig.Hearn}. The demand OD pairs in the Hearn graph are: (1,3) with 10 inflow, (1,4) with 20 inflow, (2,3) with 30 inflow, and (2,4) with 40 inflow. The results are provided in Table \ref{table.resultCTAP}. According to Table \ref{table.resultCTAP}, when converging to the same $RGAP=1e-4$, Xu \textit{et al.}'s method \cite{GG_PPATAP} has more oversaturated links than that of the CPPA-CTAP. The results generated by CPPA-CTAP are only 2\% and 1\% oversaturated on links 12 and 18, respectively. However, the results generated by Xu \textit{et al.}'s method \cite{GG_PPATAP} are approximately 3\%, 60\%, 38\%, 60\%, 14\%, and 32\% oversaturated on links 1, 3, 7, 11, 17, 18, respectively. This experiment intuitively demonstrates the effectiveness of CPPA in controlling link flow. It also proves the effectiveness of the CPPA-CTAP. However, controlling link flow comes at a price of efficiency as Xu \textit{et al.}'s method \cite{GG_PPATAP} only takes 0.72 seconds to converge while the CPPA-CTAP takes 11.06 seconds.

We also compared the proposed CPPA-CTAP with Nie's method (Nie's ALM and Nie's IPF) \cite{GG_CTAPNie}, Prashker's method \cite{prashker2004gradient} and Yang's method \cite{yang1995traffic} in the  Sioux Falls network \cite{SFnetwork}. A feasible Sioux Falls network is generated by the CUP method \cite{GG_CTAPNie}. Given the $RGAP=1e-4$, the CPPA-CTAP needs 30.10 seconds to converge while Prashker's method and Yang's method take approximately 1012.91 and 69825 seconds to converge \cite{prashker2004gradient}, respectively. The effectiveness of CPPA-CTAP has been proven. However, in a feasible Sioux Falls network, Nie's ALM and Nie's IPF only take 0.22 and 0.43 seconds to converge \cite{GG_CTAPNie}. Though the CPPA-CTAP ranks No.3 in the experiment, the running time of it is still not satisfying as the Sioux Falls network only has 24 nodes and 78 links. Thus, improving the efficiency of the CPPA-CTAP would be our future study.

\subsection{Discussion}
\label{sec.4.4}
We discuss the findings of the experiments below:
\begin{itemize}
	\item In the experiment of testing the CPPA-MF, the effectiveness of the CPPA-MF has been proven since it could generate the correct answers. It also indicates that using the CPPA to control the link flow in the PPA is effective. The CPPA-MF is much more efficient than Wang's two methods \cite{GG_PPAMC_ZWang} and is more robust than Wang's method \cite{GG_PPAMC_QWang}, which means that the superiority of the CPPA-MF to the other PPA-based MF algorithm has also been proven. We also discussed the influence of the threshold parameter $k$ on the accuracy and efficiency of the CPPA-MF.  According to our experiments, we suggest $k\in [0.7,1)$. Future work is expected for a more systematic method to determine $k$. 
	\item In the experiment of testing the CPPA-MCMF, both the efficiency and effectiveness of the CPPA-MCMF are proven. Given certain termination criteria, the CPPA-MCMF could beat the classic MCMF method (the BF algorithm \cite{BF_algorithm}) while generating correct answers. What's more, \textbf{it is the first PPA-based MCMF algorithm}. Our CPPA framework enables the flow controlling in the PPA so that using PPA to solve the MCMF problem becomes possible. We also found that the stopping criterion $\epsilon$ would influence the accuracy and the efficiency of the CPPA-MF. More efforts are in need to calculate $\epsilon$. 
	\item In the experiment of testing the CPPA-CTAP, the effectiveness of the CPPA-MCMF is proven. Compared to Xu \textit{et al.}'s method \cite{GG_PPATAP} basing on the original PPA, the CPPA-CTAP could control the link flow so that the capacity constraints in the CTAP problem would not be violated. The CPPA-CTAP is also quite efficient as it beats Prashker's method \cite{prashker2004gradient} and Yang's method \cite{yang1995traffic}. However, the CPPA-TAP is still not efficient enough compared with the state-of-the-art methods and some of the capacity constraints still would be slightly violated when applying the CPPA-CTAP. Future work is expected to improve the performance of the CPPA-CTAP.
	\item Concluding all the above discussions, the success of the CPPA-MF, the CPPA-MCMF, and the CPPA-CTAP proves the validation of using CPPA to control link flow in the PPA. It also shows what we could do using the PPA when flow control in the PPA is enabled. The proposed CPPA framework is very easy to implement and is very robust as we have already proposed three applications basing on it to show that it could be applied to three different scenarios.
	\item There is still room for the improvement of the CPPA-based algorithms' efficiency. Though the CPPA-MCMF could beat the classic methods under certain circumstances, the CPPA-MF and the CPPA-CTAP are still slower than the state-of-the-art algorithms. It is because both the original PPA and the CPPA need to solve a network Poisson Eq.(\ref{eq.4}), which results in a $O(n^3)$ time complexity. The high time complexity is an inherent defect of all the current PPA-based algorithms (no matter the CPPA-based or the original-PPA-based), unless we remodel the Physarum Polycephalum organism without using the network Poisson equation. However, regardless of the efficiency problem, we believe that the significance of the CPPA or other PPA based algorithms should be recognized as "being multi-headed makes us better at solving problems" \cite{gao2019does}. What's more, the AI technics may benefit from this self-organized and adaptive-learning bio-inspired algorithm \cite{tang2019can}. 
\end{itemize}
\section{Conclusions}
\label{sec.5}
PPA, as a newly developed BIA, has attracted lots of interest in recent years. However, few studies consider how to control the flow flowing through certain links in the PPA, which motives us to propose our CPPA framework. In this manuscript, the Capacitated Physarum Polycephalum Inspired Algorithm (CPPA) is proposed to enable capacity constraints of the link flow in the Physarum-Solver. In section \ref{sec.3.1.3}, we use a simple example to demonstrate the difference between the CPPA and the original PPA. The effectiveness of the CPPA is also initially proved in this section. To further prove the effectiveness of the CPPA and demonstrate its robustness, we proposed three applications of the CPPA, i.e., the CPPA-MF, the CPPA-MCMF, and the CPPA-CTAP in Section \ref{sec.3}. To be noticed, the CPPA-MCMF is the first PPA-based algorithm that could solve the MCMF problem. The superiority and effectiveness of them have been proven in Section \ref{sec.4}. The CPPA's sensitivity to parameter $k$ and the stopping citation $\epsilon$ is also tested. The success of the CPPA-MF, the CPPA-MCMF, and the CPPA-CTAP suggests that the proposed CPPA framework is valid and robust. Though there are still some open issues (such as choosing a proper $k$ and $\epsilon$) needed to be discussed, the significance of the proposed CPPA framework should be recognized.

\section*{Acknowledgment}
The work is partially supported by National Natural Science Foundation of China (Grant No. 61973332).

\bibliography{CPPA.bib}

\end{document}